# DSA: Evidence-Aware LLM-Agent Orchestration for Multi-Market Stock Research

**Linsen Zhu** and **Yi Shi**

## Abstract

Large language models can summarize financial information, but an operational stock-research system must first assemble heterogeneous evidence, expose unavailable data and model capabilities, and control how generated opinions affect a final report. We present DSA, an evidence-aware orchestration framework for multi-market stock research with large language model (LLM) agents. DSA organizes the workflow into evidence acquisition, structured context construction, model-routed analysis, optional role and Strategy Skill reasoning, and report generation with selected context and diagnostics. A default report profile and an optional agentic profile share evidence and model-routing services but use profile-specific output validation and risk safeguards. In the agentic profile, core role outputs are processed by role-specific parsers, whereas Strategy Skill opinions undergo an additional signal-eligibility partition before synthesis; disagreement is supplied explicitly to the decision agent, followed by a conservative risk override. The reference implementation includes six regional market paths, fifteen bundled Strategy Skills, hosted and local model routes, and multiple execution and delivery surfaces. At a frozen software snapshot, a selected manifest of 1,457 portable offline backend contract tests passed; 596 cases were retrospectively mapped to six contract families central to the reported LLM-agent architecture. This evidence establishes implementation conformance for the tested software contracts, not superior report quality, forecasting accuracy, or investment returns.

**Code:** https://github.com/ZhuLinsen/daily_stock_analysis

## 1. Introduction

Stock research is often presented to a language model as a question-answering task. In operation, the answer is the final step of a longer process. A system must identify an instrument and market, retrieve time-sensitive quotes and news from fallible sources, calculate structured indicators, select a compatible model, determine which tools and analytical roles may act, and preserve enough context to inspect the resulting judgment. Fluency alone is insufficient when unavailable data are silently treated as neutral evidence or when generated opinions acquire decision authority without an explicit control path.

Financial language models and agents have advanced domain adaptation, tool use, memory, role specialization, and collaborative decision making [1,3-7]. Their reported endpoints range from financial analysis and report generation to sequential trading and portfolio construction. Deployable research systems face an additional systems problem: markets expose unequal fields, model providers expose unequal structured-output and tool capabilities, and core role agents differ from optional analytical extensions. These asymmetries should remain observable rather than being hidden behind a uniform conversational interface.

DSA addresses this problem as an LLM-based stock-research system rather than an autonomous trading policy. Its default report profile assembles market data, technical features, fundamentals, news, and strategy evidence into a bounded context and generates a structured report. Its optional agentic profile adds role-specialized reasoning, tool calls, declarative Strategy Skills, disagreement handling, and post-synthesis risk control. The agentic path is therefore one controlled mechanism inside the wider evidence-to-report workflow, not the definition of the entire system.

This paper is a system and artifact report. It makes four contributions:

1. It presents a five-stage evidence-to-report architecture for repeatable LLM-based stock research across heterogeneous markets and data sources.

2. It separates model routing from research logic and distinguishes configured core role agents from optional Strategy Skill extensions with different output-admission mechanisms.

3. It specifies profile-level control semantics for evidence state, capability admission, output validation, risk handling, and observable failure without claiming one shared implementation for every profile.

4. It documents an open-source reference implementation and reports a frozen, claim-mapped conformance evaluation whose boundary is software behavior rather than analytical or financial effectiveness.

## 2. Related Work

### 2.1 Financial language models and tool-using agents

FinGPT develops open resources and pipelines for financial language models [1]. FinAgent combines multimodal market information, memory, and tools for trading-oriented reasoning [3], while FinMem studies layered memory and character design for a financial agent [4]. These systems establish that financial reasoning can benefit from domain evidence, memory, and executable tools. DSA adopts these components as infrastructure concerns but focuses its paper-level claim on how heterogeneous evidence and generated opinions move through an operational research workflow.

### 2.2 Specialized and collaborative financial agents

FinRobot presents a layered open-source platform for financial applications and specialized agents [5]. FinCon organizes analysts under a manager agent and studies within- and across-episode risk control for sequential financial decisions [6]. TradingAgents models a trading-firm organization with analyst, researcher, trader, risk-management, and fund-manager roles, and evaluates trading outcomes through historical simulation [7]. AlphaAgents studies role-based agents for equity selection and portfolio construction [12]. DSA likewise uses specialized roles, but it retains a non-agent default report path and evaluates implementation conformance rather than trading performance. Table 1 reports only distinctions stated in the cited papers; an unreported property is not interpreted as absent.

**Table 1. Scope of representative financial-agent systems as reported by their papers.**

| System | Primary paper endpoint | Reported agent organization | Reported evaluation scope |
|---|---|---|---|
| FinRobot [5] | Financial applications, forecasting, and report generation | Layered platform with specialized agents, model scheduling, LLMOps, and DataOps | Two demonstration applications and generated examples |
| FinCon [6] | Sequential financial decisions and portfolio management | Manager-analyst hierarchy with memory and risk-control components | Historical single-asset and portfolio experiments with returns and risk metrics |
| TradingAgents [7] | Trading decisions in a simulated trading-firm workflow | Analysts, bull/bear researchers, trader, risk-management team, and fund manager | Historical backtesting against rule-based strategies using return and risk metrics |
| DSA | Structured stock-research reports with optional agentic analysis | Default report path plus technical, intelligence, risk, decision, and optional Strategy Skill roles | Frozen, claim-mapped software-contract conformance; no analytical-superiority claim |

### 2.3 General agent frameworks and financial evaluation

ReAct couples language-model reasoning with external actions [2], while AutoGen and AgentScope provide general abstractions for multi-agent coordination and tool-enabled applications [10,11]. FinRL and FinRL-Meta provide environments for data-driven trading research [8,9]. FinToolBench evaluates financial agents against executable tool-use tasks [13], and a recent evidence review identifies time-consistent splits, execution semantics, and reproducibility as continuing bottlenecks in agentic trading studies [14]. These layers are complementary. General frameworks provide coordination primitives, and financial benchmarks test tools or policies under explicit protocols. DSA addresses the narrower control problem that appears when market evidence, model routing, core agents, optional strategy extensions, and report generation coexist in one research system. Comparative analytical evaluation remains future work.

## 3. DSA Framework

### 3.1 Research task and execution profiles

A DSA research cycle may begin with a user-specified instrument, a portfolio, or a bounded candidate set produced by the screening engine. The request supplies market and runtime settings and may select one or more Strategy Skills. The system returns a structured research report and can retain a query-linked subset of context and diagnostics for later inspection. This inspectable record is not a complete immutable event log: it does not guarantee exact replay of every provider response, model exchange, or tool call, and it does not provide claim-level provenance for every generated sentence.

DSA provides two execution profiles:

- The **default report profile** builds a bounded evidence context, invokes a configured generation model, validates the report structure, applies report-level and market-context safeguards, and renders the result. It supports scheduled and batch analysis with fewer model calls and agent dependencies.
- The **agentic profile** reuses the evidence and model-routing substrate but distributes analysis across configured role agents and optional Strategy Skills. It adds disagreement-aware synthesis and a post-synthesis risk override at the cost of additional calls, tool dependencies, and failure modes.

The default report profile is not a weaker agent, and the agentic profile is not assumed to be better. They are alternative execution paths whose report quality, latency, and cost have not yet been compared under a common protocol.

### 3.2 Five-stage evidence-to-report workflow

Figure 1 summarizes the logical workflow. The stages describe responsibilities rather than one identical function sequence for every request.

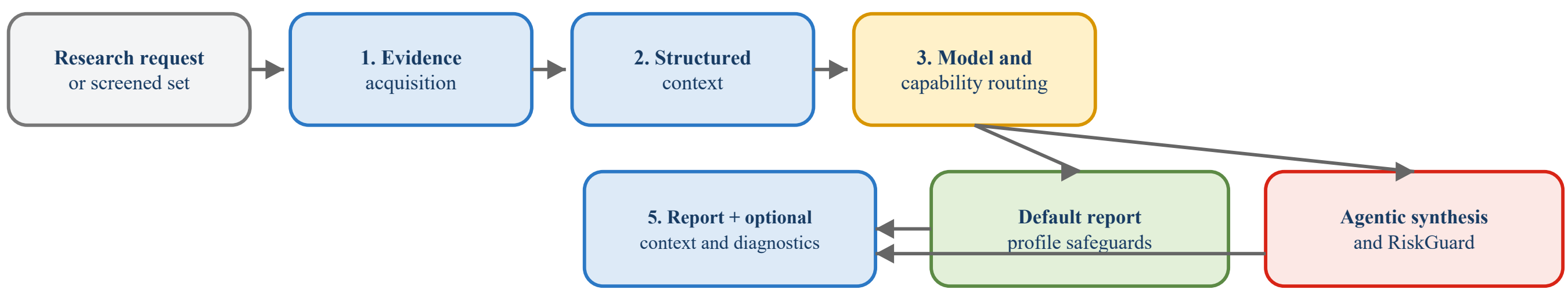

**Figure 1.** DSA evidence-to-report workflow. The two profiles reuse evidence construction and model routing, then apply different validation and risk mechanisms before producing a common report-facing result.

**Stage 1: evidence acquisition.** Market-specific adapters retrieve quotes, bars, fundamentals, announcements, news, and optional sentiment. When analysis starts from screening, deterministic filters, factor scores, risk constraints, and optional LLM reranking first produce a bounded candidate set. Instrument normalization provides a common entry surface but does not imply equal feature coverage across mainland China, Hong Kong, the United States, Japan, South Korea, and Taiwan.

**Stage 2: structured context construction.** The system calculates technical features and builds bounded evidence blocks. Availability and degradation states are normalized. Source and timestamp metadata are retained when supplied by the underlying provider or evidence block. Declarative strategies may add analytical perspectives, but their outputs remain distinguishable from observations.

**Stage 3: model and capability routing.** Provider-specific request formats and configured model names are resolved below the research workflow. Hosted services, compatible endpoints, provider-specific routes, and local models can therefore serve the same higher-level report task. Capability admission remains route-specific: a generation route and a tool-capable agent route are not assumed to be interchangeable.

**Stage 4: optional role reasoning and controlled synthesis.** Core role agents participate through configured orchestration and role-specific output parsing. Strategy Skills are optional extensions whose opinions undergo an additional signal-eligibility partition. The decision agent receives the retained role opinions, eligible Strategy Skill opinions, and an explicit disagreement summary. A subsequent agentic risk step may conservatively adjust the initial decision.

**Stage 5: report and inspection record.** Both profiles render a report for downstream clients and notification channels. When snapshot persistence is enabled, the system also stores selected context, evidence-quality summaries, and diagnostics. Presentation and transport layers do not redefine the analytical conclusion.

## 3.3 Evidence construction across unequal markets

DSA normalizes instrument identifiers and common report fields while keeping market-specific capability visible. Implemented adapters do not imply equivalent fields, historical depth, freshness, or continuous service across providers.

An evidence block may record states such as `available`, `missing`, `not_supported`, `fallback`, `stale`, `estimated`, `partial`, or `fetch_failed`. Coverage varies by block, and some upstream failures can collapse into a missing state. The distinction remains useful because unavailable capital-flow evidence should not become a neutral capital-flow signal, and a fallback should not erase the condition that triggered it. The context presented to the model therefore distinguishes retrieved evidence from at least the failure and degradation states supported by the corresponding block.

## 3.4 Model abstraction and capability admission

The generation layer supports hosted services, compatible API endpoints, provider-specific routes, and local models. The agent layer can additionally admit tool-capable backends when the required protocol is configured. The framework treats model identity as configuration rather than embedding one provider into the research logic.

Replaceability does not imply equivalence. Models differ in context capacity, structured-output behavior, tool calling, vision, latency, cost, and regional availability. DSA resolves requested operations against the relevant backend route and reports unsupported capability rather than silently substituting a semantically different operation.

## 3.5 Core role agents and Strategy Skills

The stock-level agentic workflow separates several responsibilities:

- a **technical agent** interprets price, volume, indicators, and market structure;
- an **intelligence agent** examines news, announcements, catalysts, and sentiment;
- a **risk agent** identifies uncertainty, missing evidence, and downside constraints;
- optional **specialist agents** contribute configured Strategy Skill perspectives;
- a **decision agent** synthesizes the retained evidence and opinions into a structured conclusion.

Portfolio-level agents apply related role decomposition at aggregation scope. These labels describe software responsibilities, not human-like autonomy. Core role agents are admitted by the configured orchestration and processed through role-specific parsers.

The reference artifact also bundles fifteen declarative Strategy Skills, including moving-average signals, trend following, volume breakouts, event-driven analysis, growth quality, expectation repricing, market themes, Chan theory, and wave theory. Strategy definitions declare metadata and may declare tool requirements. Unlike core roles, Strategy Skill outputs pass through a post-generation signal partition before their opinions remain in the evidence chain. Invalid Strategy Skill signals are moved to diagnostics. External strategy packages remain trusted extension inputs under the current permission model and are not claimed to execute in a complete sandbox.

## 3.6 Controlled agentic synthesis

Let $E$ denote the structured evidence context, $O_core$ the opinions produced by configured core roles after role-specific parsing, and $O_ext$ the opinions produced by Strategy Skills. The post-generation eligible extension set is

$$O_{\text{ext}}^{*} = \{\, o \text{ in } O_{\text{ext}} : \text{valid_signal}(o) \,\}.$$

Strategy selection and tool admission occur earlier through configuration and routing; the equation describes the additional output partition rather than a universal gate over all agents. The implementation then constructs a disagreement summary and an initial decision:

$$D = \text{Disagreement}(O_{\text{core}} \text{ union } O_{\text{ext}}^{*}),$$
$$d_0 = \text{Synthesize}(E, O_{\text{core}}, O_{\text{ext}}^{*}, D).$$

For the agentic profile, a later risk function may preserve or reduce the aggressiveness of the initial state:

$$d_{\text{final}} = \text{RiskGuard}(d_0, \text{risk_context}).$$

In the implemented three-state path, a veto can change `buy` to `hold`; a configured downgrade can move `buy` to `hold` or `sell`, and `hold` to `sell`. The risk override does not upgrade a decision.

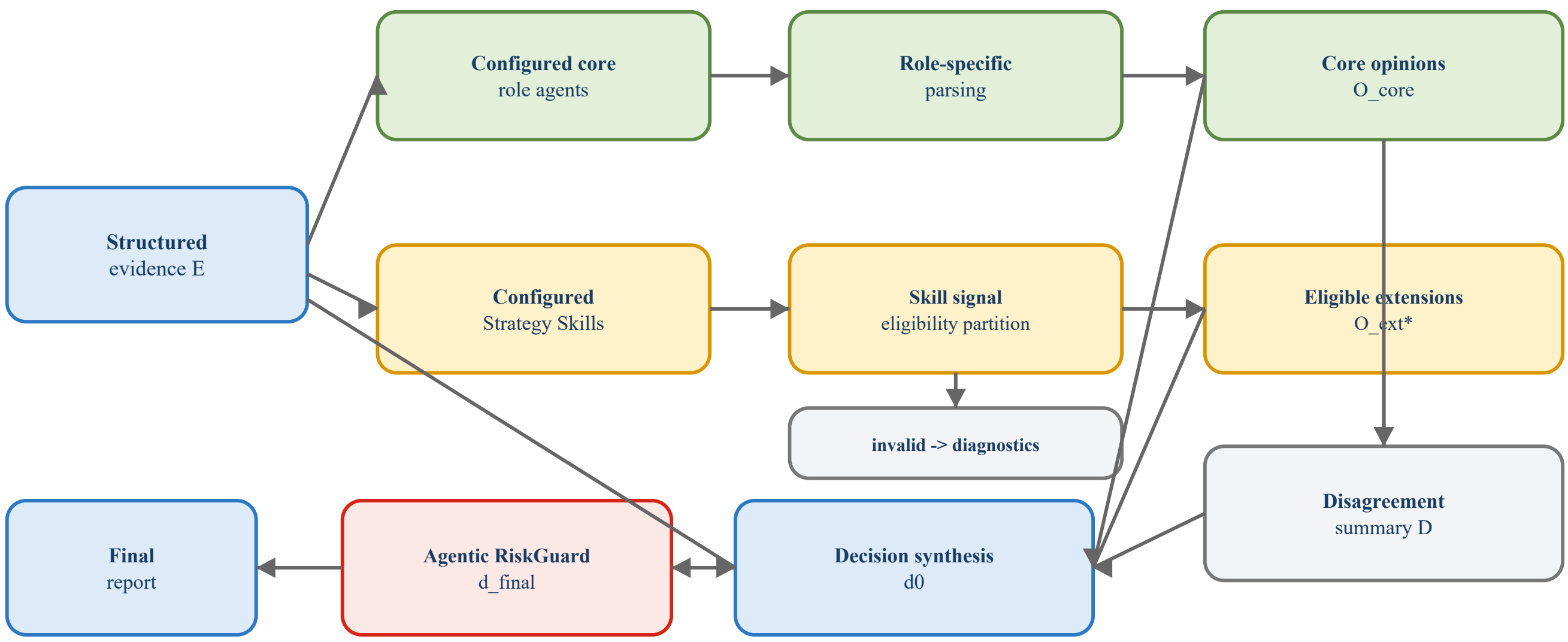

**Figure 2.** Agentic synthesis boundaries. Core role opinions and Strategy Skill opinions follow different admission paths. Only the extension path has the additional signal-eligibility partition shown here. The disagreement summary is an explicit synthesis input, and the post-synthesis RiskGuard may preserve or reduce decision aggressiveness.

Table 2 distinguishes common system intent from shared code. It does not imply one universal control component.

**Table 2. Control semantics across the two execution profiles.**

| Concern | Common substrate or intent | Default report profile | Agentic profile |
|---|---|---|---|
| Evidence | Provider adapters and bounded context construction; metadata retained when available | Evidence projected into one report-generation request | Evidence scoped across roles and synthesis |
| Capability | Configured backend and route status remain observable | Generation-route admission and report fallback | Agent-backend and tool-route admission |
| Output authority | Generated text must satisfy the consuming path's expected structure | Report parsing, schema checks, and integrity safeguards | Role-specific parsing; additional Strategy Skill signal partition before synthesis |
| Risk | Conservative intent; neither profile is an execution engine | Report-level, phase, and market-context safeguards | Post-synthesis RiskGuard with recorded veto or downgrade transitions |
| Failure and record | Partial results and diagnostics may be retained | Report/history persistence with optional selected context snapshot | Agent runtime facts, invalid Skill diagnostics, disagreement, and risk outcomes projected into the result path |

# 4. Implementation and Reproducibility

DSA is implemented primarily in Python. Evidence construction, model routing, role agents, Strategy Skills, report schemas, and rendering are organized behind explicit interfaces. The default report and agentic profiles reuse provider and model-routing services while preserving their distinct orchestration and safeguard implementations.

The reference implementation contains market paths for mainland China, Hong Kong, the United States, Japan, South Korea, and Taiwan. Its screening service incorporates an attributed and adapted implementation derived from AlphaSift [16]. Screening uses dedicated candidate and score structures; portfolio analysis likewise has dedicated snapshot and allocation structures. Batch analysis reuses the single-instrument analysis path more directly.

The artifact contains fifteen bundled Strategy Skills and multiple model routes. Command-line and API entry points, browser and desktop clients, scheduled jobs, and messaging adapters expose reports through different operational surfaces. These interfaces demonstrate that the same research workflow can be invoked and delivered in multiple environments; they are implementation scope, not separate evidence of AI quality. DSA may generate non-executable position or allocation suggestions, but the evaluated system does not calculate broker orders or place, modify, or cancel trades.

DSA-authored code is released under the MIT License. The bundled AlphaSift-derived screening components are distributed under Apache License 2.0 with attribution and modification notices recorded in the repository [15,16].

The conformance result reported below was produced from frozen Git revision `0ca56cbee2dff5cf23b1fc59c16e2d48e61ba85c` on 26 August 2026 in a Python 3.10.9 environment with pytest 7.1.2 and LiteLLM 1.89.3. The exact manifest, command, environment record, JUnit XML, checksums, and platform-boundary probe were retained as the frozen evaluation record. The public repository is a moving implementation and should not be used as a substitute for the evaluated revision and recorded environment.

# 5. Implementation Conformance Evaluation

## 5.1 Evaluation question and frozen scope

The evaluation asks whether a frozen implementation exercises the framework semantics described in Section 3. It does not ask whether DSA produces better research than a single LLM, a human analyst, or another financial-agent system, and it does not evaluate investment returns.

The authoritative portable manifest was assembled after exploratory runs exposed dependency and native-Windows boundaries, then frozen before the recorded run. It collected 1,458 cases, explicitly deselected one case tied to an unsupported native-Windows route, and executed 1,457 cases with zero failures, errors, or skips in 102.532 seconds. The manifest is therefore selected, contributor-authored, post-exploration, and claim-mapped; it is not the complete repository suite or a prospectively registered statistical sample.

Platform-specific checks were separated rather than silently counted as portable failures. A separate native-Windows boundary probe exposed platform-unsupported behavior in the Codex App Server route and Unix-shell assumptions in Docker-entrypoint tests. These outcomes delimit portability and do not support a claim that all repository tests pass on every operating system.

### 5.2 Claim-mapped contract results

Table 3 groups six mutually exclusive file families that most directly exercise the LLM-agent architecture. The remaining portable tests cover provider fallback, report integrity, multi-market APIs, interactive services, notifications, scheduling, CI, and packaging. Counts describe executed test cases, not independent observations.

**Table 3. Retrospective offline-test subset mapped to the reported LLM-agent contracts.**

| Contract family | Passed | Principal behavior exercised |
|---|---|---|
| Evidence context | 74 | context construction, evidence state, prompt projection, phase guardrails |
| Model backends, routing, and configuration | 194 | channel resolution, fallback order, capability status, generation and agent routes |
| Agent orchestration and routing | 85 | role sequencing, stock scope, tool registration, structured role outputs |
| Runtime facts, disagreement, risk, and guardrails | 87 | disagreement records, veto and downgrade transitions, runtime projections |
| Strategy registry and tool surface | 60 | discovery, metadata, required tools, configured execution admission |
| Strategy attribution and outcome lifecycle | 96 | opinion identity, attribution, outcome recording, weight boundaries |
| **Total** | **596** | **All executed cases in the mapped subset passed** |

The tests support bounded implementation conclusions. Evidence-state distinctions are exercised through context and prompt projections. Model and tool routes expose unsupported capability states in the tested paths. Core role sequencing and structured parsing are covered separately from the additional Strategy Skill validity partition. Disagreement enters decision synthesis, and the agentic risk override is tested against conservative transitions. The six-family subset preserves the frozen accounting used for this report but is not exhaustive: broader multi-agent and tool-surface cases remain in the 1,457-case record rather than in this subset. These findings do not establish the factual correctness of live provider data or model-generated financial analysis.

The tests were authored by project contributors, and many assertions exercise related code paths. Mocked services reduce environmental variance but cannot establish semantic quality under live data, models, or credentials. Dependency sensitivity was also observed: the final run used LiteLLM 1.89.3 after one prompt-cache test timed out under 1.98.0, a version that was within the repository's then-declared range.

### 5.3 Questions not answered by this evaluation

Three central empirical questions remain open. First, the default report and agentic profiles have not been compared on the same point-in-time stock-research tasks with blinded human or model-assisted rubrics. Second, role decomposition, evidence-state disclosure, Strategy Skills, disagreement handling, and risk control have not been isolated through ablation. Third, model-provider differences in report quality, latency, token use, and cost have not been measured under a common protocol.

Financial returns remain outside the current evaluation because the endpoint studied here is a research report and inspection record. DSA may emit non-executable position or allocation suggestions, but a separate downstream protocol would be required to translate them into orders, transaction costs, execution timing, and portfolio returns.

## 6. Discussion

### 6.1 DSA as an LLM-agent research system

DSA is neither only a data-reliability layer nor only a multi-agent method. Its system contribution is the integration of a repeatable stock-research workflow with an optional agent mechanism. Evidence construction and model routing provide a common substrate. The default report profile applies direct report validation and safeguards. The agentic profile adds configured core roles, optional Strategy Skills, an extension-opinion eligibility partition, disagreement-aware synthesis, and a post-synthesis risk override.

This distinction prevents multi-agent dialogue from becoming a requirement for every request and creates a natural pair of execution profiles for future quality, latency, and cost evaluation. The present evidence demonstrates tested software contracts, not that decomposition improves analysis.

### 6.2 Model heterogeneity and operational delivery

Model heterogeneity affects structured output, tool use, context limits, latency, and failure behavior. Separating routing from research logic prevents one provider's interface from defining the analytical architecture and permits different hosted, compatible, provider-specific, and local routes to serve the system. Current tests verify route and capability behavior, not semantic equivalence between models.

The Web, desktop, API, scheduled, and messaging surfaces are relevant because they exercise the same research workflow under different operational constraints. Their existence is not an independent AI contribution, and adoption indicators such as stars or downloads are not evidence of analytical validity.

### 6.3 Limitations

**Analytical validity.** The current experiments establish software conformance, not factual accuracy, forecast quality, investment performance, or superiority of multi-agent reasoning.

**Temporal validity.** Live providers, news services, and model endpoints change. Evidence-state metadata reduces ambiguity but does not create an immutable point-in-time research dataset or eliminate look-ahead leakage.

**Inspection and provenance.** Snapshot persistence is optional and retains selected rather than complete context. The current artifact does not guarantee exact replay, a full tool and model event trace, or claim-level provenance.

**Model and strategy validity.** A valid schema does not make a generated conclusion correct. Several bundled technical-analysis perspectives are disputed financial heuristics, and combining them with an LLM does not establish predictive value.

**Security and authority.** External news, model outputs, and Strategy Skill packages may contain prompt injection, poisoned content, unsafe tool requests, or secrets. Registered interfaces reduce exposure, but external packages remain trusted inputs under the current permission model.

**Generalization and reproducibility.** Market coverage is uneven, and common identifiers do not imply equal data depth. The evaluated test manifest was selected after exploratory runs and depends on a pinned environment. A versioned public snapshot is needed to reproduce the exact result independently.

### 6.4 Research outlook: governed outcome evaluation

A next step is to evaluate report quality with immutable point-in-time fixtures, shared tasks, blinded rubrics, latency and cost measurement, and ablations across the two execution profiles. Recorded downstream outcomes could then support bounded, offline revision of strategy definitions. Candidate revisions should remain versioned research artifacts and pass temporal separation, walk-forward checks, transaction-cost assumptions, parameter-sensitivity analysis, and human or policy approval before entering an active registry. AlphaEvo is one experimental direction for such offline strategy evolution [17], but it is not part of the evaluated DSA artifact and no result in this paper depends on it.

## 7. Conclusion

DSA provides an open-source reference implementation for evidence-aware LLM-agent orchestration in multi-market stock research. It connects heterogeneous evidence, model and capability routing, a default report path, optional core and Strategy Skill agents, disagreement-aware synthesis, profile-specific safeguards, and multi-surface report delivery. A frozen, selected contract-test manifest supports implementation conformance for the tested components and boundaries. Establishing analytical value requires point-in-time benchmarks, profile comparisons, ablations, report-quality evaluation, and cost-aware analysis rather than additional software-test counts alone.

## Declarations

**Author contributions.** Linsen Zhu led system conception, architecture, implementation, validation, and manuscript preparation. Yi Shi contributed technical review, code-to-manuscript consistency analysis, and manuscript revision. Both authors reviewed the final manuscript and accept responsibility for its contents.

**Code availability.** The evolving implementation and tests are available at https://github.com/ZhuLinsen/daily_stock_analysis. The evaluated revision and environment are identified in Section 4; the public repository may have advanced beyond that snapshot.

**Financial-use statement.** DSA is a research and decision-support system. It may provide non-executable research opinions and allocation suggestions, but it does not place, modify, or cancel trades in the evaluated scope and does not guarantee returns.

**AI-assisted preparation.** Generative AI tools were used for language editing and code-to-manuscript consistency checks. The authors independently verified the technical claims, references, artifact mappings, and conclusions and take full responsibility for the manuscript.